\documentclass{article} 
\usepackage{iclr2027_conference,times}

\usepackage{amsmath,amsfonts,bm}

\def\eqref#1{equation~\ref{#1}}

\def\1{\bm{1}}

\DeclareMathAlphabet{\mathsfit}{\encodingdefault}{\sfdefault}{m}{sl}
\SetMathAlphabet{\mathsfit}{bold}{\encodingdefault}{\sfdefault}{bx}{n}

\usepackage{url}
\usepackage{amsmath}
\usepackage{amssymb}
\usepackage{graphicx}
\usepackage{placeins}
\usepackage{booktabs}
\usepackage{multirow}
\usepackage{makecell}
\usepackage{xcolor}
\usepackage{colortbl}
\usepackage{trimclip}
\usepackage{algorithm}
\usepackage{algpseudocode}
\usepackage{xspace}

\usepackage{tcolorbox}
\tcbuselibrary{listings,skins}
\usepackage{hyperref}

\AddToHook{env/table/begin}{%
  \setlength{\abovecaptionskip}{0pt}%
  \setlength{\belowcaptionskip}{10pt}%
}
\definecolor{promptbg}{gray}{0.965}
\definecolor{promptframe}{gray}{0.72}
\newtcblisting{promptbox}{%
  enhanced, colback=promptbg, colframe=promptframe,
  boxrule=0.3pt, arc=1pt, left=4pt, right=4pt, top=2pt, bottom=2pt,
  listing only,
  listing options={basicstyle=\ttfamily\scriptsize, breaklines=true,
    breakindent=0pt, columns=fullflexible, keepspaces=true}%
}
\newtcblisting{actionbox}{%
  enhanced, colback=promptbg, colframe=promptframe,
  boxrule=0.3pt, arc=1pt, left=5pt, right=5pt, top=4pt, bottom=4pt,
  listing only,
  listing options={basicstyle=\ttfamily\scriptsize, breaklines=true,
    breakindent=0pt, columns=fullflexible, keepspaces=true}%
}
\definecolor{promptlabel}{RGB}{70,100,150}
\definecolor{haniehcolor}{RGB}{0,100,200}

\definecolor{cuecyan}{RGB}{66,196,188}
\definecolor{cuepink}{RGB}{235,91,163}
\definecolor{cuepurple}{RGB}{116,76,224}
\definecolor{cuegray}{RGB}{242,242,244}
\newcommand{\promptrole}[1]{{\footnotesize\sffamily\bfseries\color{promptlabel}#1}\par\vspace{1pt}}

\newcommand{\method}{\textsc{CueKFS}\xspace}

\newcommand{\CD}{Cue Decomposition\xspace}
\newcommand{\CWGR}{Cue-Wave-Guided Refinement\xspace}
\newcommand{\CBA}{Cue Budget Allocation\xspace}
\newcommand{\bestf}[1]{\textbf{#1}}

\newcommand{\changemark}{\textcolor{cuepink!80!black}{\raisebox{0.15ex}{\rotatebox[origin=c]{180}{$\Lsh$}}}}

\title{\method: Agentic Cue-Driven Keyframe\\Selection for Long Video Understanding}

\author{
Weitai Kang\textsuperscript{1},
Hanieh Deilamsalehy\textsuperscript{2},
Yumo Xu\textsuperscript{2},
Dewang Sultania\textsuperscript{2},
Serdar Cellat\textsuperscript{2},
Yan Yan\textsuperscript{1}
\\
\textsuperscript{1}University of Illinois Chicago
\quad
\textsuperscript{2}Netflix
}

\iclrfinalcopy
\begin{document}
\maketitle

\begin{abstract}

Keyframe selection (KFS) has long produced compact video summaries for browsing and retrieval, and
representative frames for thumbnails. 
More recently, when conditioned on a question, KFS provides an alternative to uniform sampling for long-video question answering by selecting frames that are more relevant to the question.
Most methods rank frames by similarity to the question. Yet a relevant frame may score poorly when
the question combines subjects or moments that no single frame shows, or requires implicit information absent
from its wording. Other methods try to break down the question into subqueries, but suffer from inaccurate decomposition due to their static initial context.
Therefore, we propose \method, a training-free method that reformulates question--frame matching as comparing frames against a set of dynamically generated visual cues. From an initial set of salient frames, we decompose the question into cues. Each cue concurrently probes the video to navigate to its own evidence. A reasoning VLM then agentically revises the cue set against its evidence to re-explore the video. \method\ then allocates the budget across the surviving cues. 
Across three benchmarks, \method\ establishes state-of-the-art results in all $27$ evaluated settings with available prior results, achieving budget-averaged gains of up to +4.54\% over the previous baseline and a median of only two VLM calls. We further provide a detailed behavioral analysis of \method, showing that agentic cue refinement drives active re-exploration of the video, yielding relative similarity gains of up to 92\% over the initial context.
\end{abstract}

\section{Introduction}

Keyframe selection (KFS), previously studied as video summarization, picks a small set of representative frames from a video and has long been valuable in summarization, retrieval, and thumbnail
selection~\citep{gygli2014,clipit,vssurvey2025}. 
More recently, KFS has been adopted as an upstream alternative to uniform sampling in video question answering. By tailoring the input frames to the question, it increases their relevance with the same number of frames or reduces this number when possible~\citep{aks2025,bolt2025,wfssb}.

Most methods treat KFS as question--frame matching: they score frames by embedding similarity to the question and select from that  signal~\citep{clip,siglip2,aks2025,mdp3,bolt2025,qframe,wfssb,keyvideollm}. 
This primarily rewards surface similarity, yet such similarity alone is insufficient to determine frame relevance: a question may span several subjects or moments that no single frame captures in full, concern causal or temporal relations that no single frame can show, or rely on implicit information, such as genre or intent, that the question’s wording does not name.
Other methods decompose the question into parts~\citep{msjoe,himu2026,toolmerge2026,lif2025,tstar} from the question alone or a coarse video preview, leaving much of the video unseen before retrieval targets are defined. Therefore, subsequent retrieval may refine the regions to examine but remains constrained by a potentially biased initial decomposition.

An embedding model can score each frame efficiently but has limited reasoning capacity. Instead, a VLM offers stronger reasoning but is costly to apply across an entire video, as in dense captioning. 
We therefore propose \method, a training-free KFS method that uses the embedding model for video-wide retrieval and the reasoning VLM to dynamically revise what should be retrieved next. 
Specifically, a VLM first decomposes the question into visual cues based on a compact video overview, each cue a short self-contained phrase describing one concrete visual element. The embedding model retrieves evidence for each cue across the video, and the VLM agentically revises the cues guided by that evidence. The revised cues then perform another video-wide retrieval, after which \method\ allocates the $k$-frame budget across the surviving cues.
We design \emph{peaks} and \emph{plateaus} to surface that evidence, and \emph{Batched Agentic Refinement} to revise the cues efficiently.

Our contributions are threefold. (i) We propose \method, a method that turns a reasoning VLM with an embedding model into an agentic explorer of long video, matching dynamic cues against all frames to retrieve evidence. 
(ii) Across three benchmarks, three downstream VLMs, and three frame budgets, \method\ establishes state-of-the-art results in all 27 settings with available prior results, achieving budget-averaged gains of up to +4.54\% over the previous baseline and a median of only $2$ reasoning-VLM calls per selection.
(iii) We further provide behavioral analysis of our method, showing that \method actively explores the video beyond its initial context: on long Video-MME videos, final cues achieve up to a 92\% higher match score on frames outside the initial context.

\section{Related Work}
\label{sec:related}

\subsection{Selection by question relevance}
Most methods score each candidate frame by its similarity to the whole
question and add structure to that single signal: \textsc{AKS} balances relevance and
coverage~\citep{aks2025}, \textsc{MDP$^3$} adds diversity and sequentiality~\citep{mdp3}, \textsc{BOLT} applies inverse transform sampling over the similarity distribution~\citep{bolt2025},
\textsc{Q-Frame} adds multi-resolution adaptation~\citep{qframe}, and \textsc{WFS-SB}~\citep{wfssb} denoises the signal to detect semantic boundaries. Others keep the single question-level signal but condition it, routing by a coarse query type (\textsc{DIG}~\citep{dig2026}) or weighting fixed modality streams by query intent (\textsc{Q-Gate}~\citep{qgate2026}). \textsc{A.I.R.}~\citep{air2026} keeps the same signal but adds a VLM that verifies its top-ranked frames and densifies the sampling around the ones it confirms. 
Most of these rank a frame by a single question-level score, an imperfect proxy for its relevance. 
Some methods also use question--frame similarity in data construction, training, or scoring~\citep{framevoyager,frameoracle,gens,evidential2026}, and are thus contaminated by this proxy.
\method\ instead decouples the question into cues and scores each frame against every cue, so a frame relevant to any one cue still surfaces.

\subsection{Selection by question decomposition}
\label{sec:related-rewrite}
Other methods decompose the question into multiple parts before matching. \textsc{MSJoE} reasons out several queries from a coarse preview but merges them into
one similarity matrix consumed by a trained sampler~\citep{msjoe}, \textsc{HiMu} parses the question text alone into a hierarchical logic tree of atomic predicates composed into one satisfaction
curve~\citep{himu2026}, and \textsc{ToolMerge} decomposes it into tool calls merged into one
ranking~\citep{toolmerge2026}.
\textsc{LIF}~\citep{lif2025} extracts objects and typed relations for search with a frozen detector. \textsc{T*}~\citep{tstar} decomposes it into target and cue objects from a coarse glance, then iteratively refines where a frozen detector searches for them, though found targets can be removed.
In all of these, target descriptions are formed from the initial decomposition. Subsequent search may update where to look or remove found targets, but does not update the targets. \method\ instead lets each cue evolve against its own retrieved evidence, so the cue set explores the video dynamically rather than remaining bounded by the initial decomposition.

\subsection{Evaluating keyframe selection}
\label{sec:related-eval}
Keyframe selection is evaluated indirectly: the selected frames are supplied to a downstream VLM, and its QA accuracy on video benchmarks (Video-MME~\citep{videomme}, MLVU~\citep{mlvu}, LongVideoBench~\citep{longvideobench}) is reported as the selection's score. Studies adopt common downstream VLMs with modest capacity for comparability, but the score then conflates selection quality with the VLM's own capability: a weak VLM under-rewards a correct selection when it cannot answer even from sufficient frames. 
We retain this protocol but add a strong downstream VLM (GPT-5.5) to reduce the influence of the downstream VLM's limitations on our evaluation, and report accuracy averaged across frame budgets: small budgets keep even a strong VLM sensitive to missing or redundant frames, while averaging reduces dependence on any single budget.

\section{Method}
\label{sec:method}

\begin{figure*}[t]
    \centering
    \includegraphics[width=\textwidth]{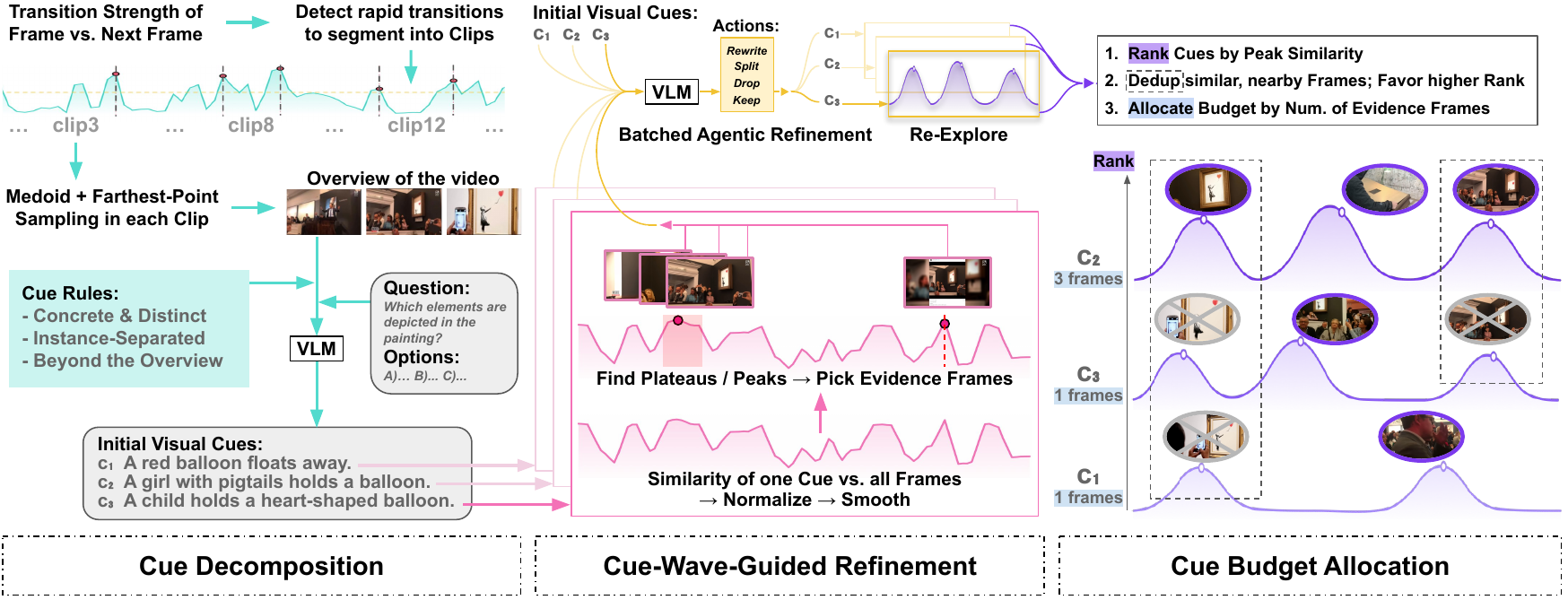}
    \caption{Overview of \method. \CD builds a clip-wise video overview and applies three rules to produce initial visual cues. \CWGR computes one wave per cue, samples evidence from plateaus and peaks, refines cues in batches with a reasoning VLM, and re-explores the video. \CBA ranks cues by peak similarity, removes similar nearby frames in favor of higher-ranked cues, and allocates the budget by remaining evidence-frame counts. The displayed cue set and frame counts are illustrative.}
    \label{fig:method-overview}
\end{figure*}

Given a video $V$, a question (or more generally, a task context) $q$, and a frame budget $k$, \method\ returns at most $k$ frames for a downstream VLM. We sample $T$ candidate frames, $\mathcal{F}=\{f_1,\ldots,f_T\}$, indexed in temporal order, and use a frozen embedding model. Its image encoder gives the frame embeddings $\mathbf{e}_t=\phi(f_t)$, while its text encoder embeds a textual cue $c$ as $\psi(c)$. As illustrated in Figure~\ref{fig:method-overview}, the method comprises \CD\ (\S\ref{sec:decompose}), \CWGR\ (\S\ref{sec:refine}), and \CBA\ (\S\ref{sec:allocate}). Section~\ref{sec:setup} and Table~\ref{tab:hp} specify the experimental configurations.

\subsection{Cue Decomposition}
\label{sec:decompose}

As shown in the left of Figure~\ref{fig:method-overview}, rapid inter-frame transitions divide the video into clips, clip representatives form a compact overview, and a VLM decomposes the question into initial visual cues.

\paragraph{Overview of the video.} We first form the question-agnostic inter-frame signal
\begin{equation}
a_t=\cos(\mathbf{e}_t,\mathbf{e}_{t+1}),\qquad t=1,\ldots,T-1.
\end{equation}
Similar to \citet{wfssb}, our segmentation uses a wavelet transform, but applies it to adjacent-frame similarity $a$ rather than question-conditioned frame relevance. Peaks in the resulting transition-strength signal define clip boundaries. This avoids propagating errors from matching the whole question directly to individual frames into the overview. Within each clip, we first select the medoid and then use it to start farthest-point sampling, yielding complementary representatives for the overview $R$ under a frame budget $N_{\text{repr}}$. Appendix~\ref{app:impl} gives the transition and sampling details.

\paragraph{Initial visual cues.} A reasoning VLM receives $R$, $q$, and candidate answers when available, together with cue-construction rules.
As shown in Figure~\ref{fig:method-overview} (left), the rules cover three principles: \emph{Concrete and Distinct}, \emph{Instance-Separated}, and \emph{Beyond the Overview}. First, each cue should be a short, self-contained description of a concrete visual element that a single frame could show, be matchable by appearance, and remain distinct from the other cues. Second, distinct instances, events, or stages in counting, ordering, before--after, and change-over-time questions are written as separate cues, making each target independently retrievable. Third, the VLM may formulate question-relevant cues even when the related frames are absent from the overview, so $R$ conditions but does not bound what the cues may seek in the initial overview. The VLM then returns the initial visual cue set $\mathcal{C}^{0}=\{c_1,\ldots,c_m\}$. Appendix Figure~\ref{fig:prompt-decompose} gives the cue-construction prompt.

\subsection{Cue-Wave-Guided Refinement}
\label{sec:refine}

As shown in the middle of Figure~\ref{fig:method-overview}, the initial cues probe all the frames in parallel, evidence frames sampled from each cue's plateaus and peaks guide batched agentic refinement, and the refined cues re-explore the full video.

\paragraph{Per-cue similarity waves.} For every initial cue $c\in\mathcal{C}^{0}$, we compare it with every candidate frame to obtain a raw similarity signal, then robustly normalize and smooth the signal into a cue wave:
\begin{equation}
s^c_t=\cos\!\big(\mathbf{e}_t,\psi(c)\big),\qquad
w^c_t=\operatorname{smooth}_{h}\!\left(
\frac{s^c_t-\operatorname{med}(s^c)}
{\operatorname{p}_{95}(s^c)-\operatorname{p}_{5}(s^c)+\varepsilon}
\right).
\label{eq:cue-wave}
\end{equation}
Here $\operatorname{med}$ and $\operatorname{p}_r$ denote the median and the $r$th percentile, and $\operatorname{smooth}_h$ is a moving average of width $h$. $\varepsilon>0$ prevents division by zero. Each $w^c$ is normalized separately to locate evidence within its cue; raw $s^c$ values are retained for cross-cue comparison in \CBA.

\paragraph{Plateaus, peaks, and evidence frames.} Each cue wave provides two complementary forms of evidence. Plateaus capture sustained matches, such as an extended scene, whereas peaks capture brief, isolated matches, such as a momentary action. We first identify maximal above-threshold intervals $[a,b]$ satisfying
\begin{equation}
w^c_u\ge Q_{\rho}(w^c)\quad\forall u\in[a,b],
\label{eq:plateau}
\end{equation}
where $Q_{\rho}(w^c)$ is the $\rho$th percentile of that cue's wave. After merging nearby intervals and discarding those shorter than $\ell_0$ frames, we select each plateau's medoid and use it to start farthest-point sampling within the interval, yielding up to $n_{\text{pl}}$ plateau samples. A peak candidate is a local maximum $t$ satisfying $w^c_{t-1}<w^c_t\ge w^c_{t+1}$. Its prominence measures its rise above its local surroundings:
\begin{equation}
\pi^c_t=w^c_t-\max\!\left(
\min_{u\in[t-\delta,t]}w^c_u,\ 
\min_{u\in[t,t+\delta]}w^c_u
\right).
\label{eq:peak-prominence}
\end{equation}
We first suppress candidates separated by fewer than $d_p$ frames in descending peak height, discard the remaining candidates below the prominence threshold $\pi_0$, and retain at most $n_p$ peaks with the highest wave values. Each retained peak contributes its center frame. The retained peak centers and plateau samples together form the cue-specific evidence set $\mathcal{E}_c$. We assign each evidence frame a score: $w^c_t$ for a peak center and the plateau's mean wave value for every plateau sample. These evidence scores prioritize frames for refinement and later rank frames within each cue; the selected refinement images are presented in temporal order within each cue. Table~\ref{tab:hp} and Appendix~\ref{app:impl} give the thresholds and sampling settings.


\begin{figure}[t]
  \centering
  \begin{minipage}{\textwidth} 
    \begin{actionbox}
    For EACH cue, choose one action:
    - drop:    the cue cannot help with the context.
    - keep:    the cue is useful for the context; keep it unchanged.
    - rewrite: the cue is useful but imprecise or retrieving the wrong thing; give
               one better cue, still faithful to the context.
    - split:   the cue bundles two unrelated visual things; give two or more
               specific cues.
    \end{actionbox}
  \end{minipage}
  \caption{Evidence-grounded cue-refinement actions. For each cue, the reasoning VLM chooses one of four actions based on that cue's wave-retrieved evidence.}
  \label{fig:refine-actions}
\end{figure}

\paragraph{Batched agentic refinement.} Since initial cues are formed before they explore the video, they may be imprecise, bundle distinct targets, or contribute no useful evidence. Therefore, we send each cue and its evidence $\mathcal{E}_c$ obtained from the previous exploration to a reasoning VLM to produce a more accurate cue for further search. Reviewing cues individually would multiply expensive VLM calls, so we process them in batches of up to $B$ cues sharing an evidence-image budget of $M$,
pairing each cue with its own evidence frames. For each cue within a batch, the VLM independently keeps it unchanged when useful, rewrites it when imprecise or retrieving the wrong thing, splits it when it bundles distinct targets, or drops it when it provides no useful evidence. Figure~\ref{fig:refine-actions} shows the corresponding prompt. Kept cues and the outputs of rewrite and split form the refined cue set $\mathcal{C}^{1}$, while dropped cues are excluded. Appendix Figure~\ref{fig:prompt-refine} gives the exact batched refinement prompt.

\paragraph{Re-explore.} 
Every rewritten or split cue in $\mathcal{C}^{1}$ re-explores all $T$ candidate frames and follows the same wave construction and evidence extraction. Kept cues retain their existing waves and evidence. The resulting evidence frames are candidates for \CBA, retrieved by dynamically refined cues rather than a static context.

\subsection{Cue Budget Allocation}
\label{sec:allocate}

As shown in the right panel of Figure~\ref{fig:method-overview}, \CBA\ ranks the refined cues by peak similarity, removes frames only when they are visually similar and temporally nearby, and allocates the frame budget according to each cue's remaining evidence count.

\paragraph{Rank cues by peak similarity.} For each refined cue $c\in\mathcal{C}^{1}$, we define its peak similarity as
\begin{equation}
\iota(c)=\max_{1\le t\le T}s^c_t.
\label{eq:cue-importance}
\end{equation}
Unlike the separately normalized waves $w^c$, the raw values $s^c$ are comparable across cues. Sorting by $\iota(c)$ resolves shared evidence, prioritizes cues when the budget cannot cover them all, and breaks integer-allocation ties; proportional targets depend on evidence counts.

\paragraph{Deduplicate similar, nearby frames.} To avoid spending multiple slots on the same evidence, we process all retrieved evidence frames from higher to lower cue rank, ordering frames within each cue by evidence score. Two evidence frames $f_i$ and $f_j$, occurring at times $t_i$ and $t_j$, are considered redundant only when
\begin{equation}
\cos(\mathbf{e}_i,\mathbf{e}_j)\ge\tau_{\cos}
\quad\text{and}\quad
|t_i-t_j|\le\tau_{\Delta}.
\label{eq:dedup}
\end{equation}
Requiring both conditions preserves repeated-looking events at distant times and distinct events that occur close together. A frame is retained only if it is not redundant with any previously retained frame. When redundant frames come from different cues, the frame retrieved by the higher-ranked cue is retained. This leaves no redundant pairs among the evidence frames passed to allocation.

\paragraph{Allocate budget by evidence count.} After de-duplication, we divide the frame budget $k$ among cues in proportion to the number of evidence frames each retains. When the budget permits, we first assign one slot to each cue to preserve coverage, then assign each remaining slot to the cue with the largest shortfall from its proportional target for the total budget, subject to its available evidence count. If cue $c$ receives $k_c$ slots, we select its $k_c$ highest-scoring retained evidence frames. Appendix~\ref{app:impl} gives more details. The final frames are returned in temporal order.

\section{Experiments}

\subsection{Experimental setup}
\label{sec:setup}

\paragraph{Evaluation configuration.}
Following the standard evaluation protocol, a KFS method receives a video and a multiple-choice question, and returns $k$ frames. A frozen downstream VLM then answers the same question using only the selected frames. We report accuracy on Video-MME~\citep{videomme}, MLVU~\citep{mlvu}, and LongVideoBench~\citep{longvideobench}, without subtitles. Candidate frames are sampled at $1$\,fps with budgets $k\in\{8,16,32\}$. To reduce sensitivity to the choice of frame budget, we also report the mean across the three budgets, denoted by ``Avg.'' Since VideoQA accuracy depends on both the selected frames and the downstream VLM (\S\ref{sec:related-eval}), we use GPT-5.5~\citep{gpt55} as a downstream VLM. Table~\ref{tab:main} also reports results with Qwen2.5-VL-7B~\citep{qwen25vl} and InternVL3-8B~\citep{internvl3} for comparison with prior KFS work.


\paragraph{CueKFS configuration.}
We use the frozen SigLIP2-giant embedding model~\citep{siglip2}. For reproducibility, we use the open source Qwen3.6-27B~\citep{qwen36} as the reasoning VLM for the ablation studies in Table~\ref{tab:ablation} and the analysis of cue behavior in Section~\ref{sec:analysis}. The full \method\ configuration for these experiments has $24$ overview frames, batches of $5$ cues sharing $24$ evidence frames, and $(\tau_{\cos},\tau_\Delta)=(0.95,2\text{s})$. For the results in Table~\ref{tab:main}, we use GPT-5.5 as the reasoning VLM. Since \method\ has several configurable stages, Table~\ref{tab:hp} examines how the reasoning VLM and hyperparameter settings affect performance and reports the configuration used in Table~\ref{tab:main}. Appendix~\ref{app:impl} provides additional implementation details.

\subsection{Main results}
\label{sec:main}

\begin{table}[t]
\caption{Accuracy (\%) on three long-video QA benchmarks at three frame budgets with three downstream VLMs. Bold denotes the best result in each block. For Qwen2.5-VL and InternVL3, baseline results at $k{=}32$ are from \citet{wfssb}.}
\label{tab:main}
\centering
\small
\setlength{\tabcolsep}{3.2pt}
\renewcommand{\arraystretch}{1.08}
\resizebox{0.98\textwidth}{!}{%
\begin{clipbox}{0pt 0pt 0pt 0pt}
\begin{tabular}{@{}ll cccc cccc cccc@{}}
\toprule
\multirow{2}{*}{\makecell{Downstream\\VLM}} &
\multirow{2}{*}{KFS method} &
\multicolumn{4}{c}{Video-MME} & \multicolumn{4}{c}{MLVU} & \multicolumn{4}{c}{LongVideoBench} \\
\cmidrule(lr){3-6}\cmidrule(lr){7-10}\cmidrule(lr){11-14}
&
& $k{=}8$ & $k{=}16$ & $k{=}32$ & Avg
& $k{=}8$ & $k{=}16$ & $k{=}32$ & Avg
& $k{=}8$ & $k{=}16$ & $k{=}32$ & Avg \\
\midrule
\multirow{3}{*}{GPT-5.5}
 & Uniform & 73.70 & 77.11 & 81.41 & 77.41 & 61.73 & 67.74 & 71.90 & 67.12 & 63.95 & 67.46 & 70.38 & 67.26 \\
 & WFS-SB~\citep{wfssb} & 76.22 & 78.74 & 80.96 & 78.64 & 76.13 & 80.36 & 82.11 & 79.53 & 69.78 & 73.90 & 76.66 & 73.45 \\
\rowcolor{cuepurple!11}
 & \method & \bestf{80.67} & \bestf{83.30} & \bestf{84.15} & \bestf{82.70} & \bestf{80.08} & \bestf{82.29} & \bestf{83.49} & \bestf{81.95} & \bestf{75.39} & \bestf{77.26} & \bestf{79.88} & \bestf{77.51} \\
\midrule
\multirow{6}{*}{Qwen2.5-VL}
 & Uniform & 53.04 & 57.56 & 60.00 & 56.87 & 54.14 & 58.37 & 61.13 & 57.88 & 52.80 & 56.32 & 58.71 & 55.94 \\
 & AKS~\citep{aks2025} & -- & -- & 64.0 & -- & -- & -- & 67.2 & -- & -- & -- & 63.2 & -- \\
 & MDP$^3$~\citep{mdp3} & -- & -- & 63.8 & -- & -- & -- & 66.2 & -- & -- & -- & 60.0 & -- \\
 & A.I.R.~\citep{air2026} & -- & -- & 65.0 & -- & -- & -- & 67.5 & -- & -- & -- & 61.4 & -- \\
 & WFS-SB~\citep{wfssb} & 58.37 & 61.30 & 64.40 & 61.36 & 66.05 & 68.35 & 70.40 & 68.27 & 59.91 & 63.35 & 64.40 & 62.55 \\
\rowcolor{cuepurple!11}
 & \method & \bestf{64.85} & \bestf{66.26} & \bestf{66.59} & \bestf{65.90} & \bestf{69.55} & \bestf{71.57} & \bestf{71.48} & \bestf{70.87} & \bestf{62.30} & \bestf{64.70} & \bestf{65.00} & \bestf{64.00} \\
\midrule
\multirow{6}{*}{InternVL3}
 & Uniform & 59.41 & 62.56 & 64.44 & 62.14 & 64.17 & 67.07 & 69.23 & 66.82 & 52.95 & 56.25 & 59.01 & 56.07 \\
 & AKS~\citep{aks2025} & -- & -- & 66.3 & -- & -- & -- & 74.2 & -- & -- & -- & 61.5 & -- \\
 & MDP$^3$~\citep{mdp3} & -- & -- & 66.8 & -- & -- & -- & 74.0 & -- & -- & -- & 60.9 & -- \\
 & A.I.R.~\citep{air2026} & -- & -- & 68.2 & -- & -- & -- & 74.5 & -- & -- & -- & 62.8 & -- \\
 & WFS-SB~\citep{wfssb} & 64.37 & 65.48 & 67.40 & 65.75 & 73.64 & 75.62 & 74.80 & 74.69 & 59.76 & 60.81 & 62.90 & 61.16 \\
\rowcolor{cuepurple!11}
 & \method & \bestf{68.11} & \bestf{69.26} & \bestf{69.19} & \bestf{68.85} & \bestf{76.54} & \bestf{77.14} & \bestf{77.64} & \bestf{77.11} & \bestf{62.83} & \bestf{64.17} & \bestf{64.62} & \bestf{63.87} \\
\bottomrule
\end{tabular}%
\end{clipbox}
}
\end{table}

As shown in Table~\ref{tab:main}, \method\ achieves state-of-the-art accuracy in all $27$ settings. With GPT-5.5 as the downstream VLM, its budget-average gains over WFS-SB~\citep{wfssb} on Video-MME, MLVU, and LongVideoBench are $+4.06$, $+2.42$, and $+4.06$. The corresponding gains over WFS-SB are $+4.54$, $+2.60$, and $+1.45$ with Qwen2.5-VL, and $+3.10$, $+2.42$, and $+2.71$ with InternVL3. With GPT-5.5, \method\ also beats uniform sampling by $+5.29$, $+14.83$, and $+10.25$. 
At $k{=}32$, \method\ outperforms AKS~\citep{aks2025}, MDP$^3$~\citep{mdp3}, and A.I.R.~\citep{air2026} on every benchmark.

\subsection{Ablation studies}
\label{sec:ablation}

\begin{table}[t]
\caption{Step-by-step ablations on Video-MME with GPT-5.5 as the downstream VLM.}
\label{tab:ablation}
\centering
\footnotesize
\renewcommand{\arraystretch}{0.92}
\setlength{\tabcolsep}{3.2pt}
\begin{clipbox}{0pt 0pt 0pt 0pt}
\begin{tabular}{@{}p{0.6\textwidth} rrrrr@{}}
\toprule
Configuration & $k{=}8$ & $k{=}16$ & $k{=}32$ & Avg & \makecell{VLM calls\\(median)} \\
\midrule
Uniform sampling & 73.70 & 77.11 & 81.41 & 77.41 & 0 \\
\textcolor{cuecyan!75!black}{Add} Overview segmentation with per-clip question matching & 75.59 & 77.26 & 80.63 & 77.83 & 0 \\
\textcolor{cuecyan!75!black}{Add} \CD \& \CBA & 76.33 & 79.56 & 81.89 & 79.26 & 1 \\
\rowcolor{cuegray}
\qquad ~\changemark\ \textcolor{blue}{Remove} \CBA & 71.44 & 74.81 & 78.44 & 74.90 & 1 \\
\addlinespace[2pt]
\textcolor{cuecyan!75!black}{Add} Per-cue refinement using cue-wave evidence & 78.63 & 81.19 & 83.04 & 80.95 & \textcolor{red}{11} \\
\rowcolor{cuepurple!11} ~\changemark\ \textcolor{cuecyan!75!black}{Change to} Batched refinement (\CWGR) & \bestf{79.30} & \bestf{80.63} & \bestf{83.19} & \bestf{81.04} &  \changemark\bestf{2} \\
\rowcolor{cuegray}
\qquad ~\changemark\ \textcolor{blue}{Remove} \CBA & 74.85 & 79.50 & 81.63 & 78.66 & 2 \\
\rowcolor{cuegray}
\qquad ~\changemark\ \textcolor{blue}{Remove} cue-wave evidence & 75.81 & 77.85 & 80.63 & 78.10 & 2 \\
\bottomrule
\end{tabular}%
\end{clipbox}
\end{table}

Table~\ref{tab:ablation} builds \method\ step by step. We replace uniform sampling with overview segmentation, selecting the frame with the highest question--frame similarity within each clip. 
This segmentation baseline raises the average accuracy by $0.42$ points.
Adding \CD\ and \CBA\ brings a further $1.43$ point gain. Removing \CBA\ from this stage instead lowers the average by $4.36$ points, demonstrating the importance of combining decomposition with allocation.
We then add per-cue refinement grounded in cue-wave evidence, reviewing one cue per call with an average of $35.1$ evidence frames. This raises the average from $79.26$ to $80.95$, a gain of $1.69$ points, but increases the median number of VLM calls to $11$.
In the batched-refinement ablation, \CWGR\ uses a batch size of $B{=}5$ and shows an average of $5.6$ evidence frames per cue, keeping the visual context per call roughly comparable. It maintains $81.04$ accuracy while reducing median VLM calls from $11$ to $2$. 
We use batched refinement in the remaining experiments.
Removing \CBA\ from the batched-refinement configuration in Table~\ref{tab:ablation} reduces $2.38$ points, while removing cue-wave evidence reduces it by $2.94$ points. These controlled comparisons further validate both \CBA\ and the plateau-and-peak evidence used in \CWGR.

\begin{table}[t]
\caption{Hyperparameter ablations on Video-MME. Values and their corresponding average accuracies are listed in the same order. Bold marks the configuration used in Table~\ref{tab:main} and its accuracy.}
\label{tab:hp}
\centering
\footnotesize
\setlength{\tabcolsep}{8pt}
\renewcommand{\arraystretch}{0.92}
\begin{tabular}{@{}l c c@{}}
\toprule
Hyperparameter & Values evaluated & Avg \\
\midrule
Reasoning VLM
& Qwen3.6-27B / \bestf{GPT-5.5}
& 81.04 / \bestf{81.52} \\
\addlinespace[2pt]

Overview frames $N_{\text{repr}}$
& 16 / 24 / 32 / \bestf{48} / 64
& 80.50 / 81.52 / 81.32 / \bestf{82.35} / 82.35 \\

Cues per batch $B$
& 3 / 5 / 10 / \bestf{20} / all
& 82.06 / 82.35 / 82.30 / \bestf{82.47} / 82.42 \\

Review image budget $M$
& 16 / 24 / \bestf{32} / 48 / 64
& 82.42 / 82.47 / \bestf{82.55} / 82.20 / 82.13 \\

\makecell[l]{VLM frame\\edge (px)}
& \bestf{448} / 768 / 1024 / 1536
& \bestf{82.55} / 81.75 / 81.95 / 82.38 \\

Wave smoothing window $h$
& 1 / \bestf{3} / 5
& 82.29 / \bestf{82.55} / 81.29 \\

\makecell[l]{Peak prominence and\\plateau percentile}
& $(0.15,70)$ / $\mathbf{(0.25,80)}$ / $(0.35,90)$
& 82.21 / \bestf{82.55} / 82.38 \\

\makecell[l]{Deduplication thresholds\\$(\tau_{\cos},\tau_\Delta\,\text{s})$}
& \makecell[c]{$(0.95,2)$ / $(0.90,1)$ / $(0.90,3)$\\
$\mathbf{(0.98,1)}$ / $(0.98,3)$}
& \makecell[c]{82.55 / 82.35 / 82.27\\
\bestf{82.70} / 82.67} \\
\bottomrule
\end{tabular}
\end{table}

Table~\ref{tab:hp} examines how the reasoning VLM and hyperparameters affect Video-MME performance. The pipeline has multiple stages and hyperparameters, making exhaustive evaluation of their joint configuration space impractical. 
We therefore first examine the reasoning VLM and then vary one factor at a time across the three pipeline stages, from \CD to \CWGR and finally \CBA. The configuration used in Table~\ref{tab:main} uses GPT-5.5 as the reasoning VLM, $N_{\text{repr}}=48$, $B=20$, $M=32$, a $448$\,px maximum image edge for the reasoning VLM, $h=3$, $(\pi_0,\rho)=(0.25,80)$, and $(\tau_{\cos},\tau_\Delta)=(0.98,1\,\text{s})$. It yields $82.70$ average accuracy on Video-MME with GPT-5.5 as the downstream VLM.


\subsection{Analysis of cue behavior}
\label{sec:analysis}

\paragraph{Refinement is active and most effective at low budgets.}
Table~\ref{tab:refine-behavior}(a) shows that refinement rewrites at least one cue on $73\%$ of questions, drops one on $71\%$, and splits one on $33\%$. All three actions become more frequent as video length increases. This trend suggests that the initial context is less reliable for longer videos, making refinement more important for re-exploration.
Panel (b) shows positive refinement gains in all three action groups at both reported budgets. The gains are consistently larger at $k{=}8$ than at $k{=}32$, showing that refinement is especially important when the frame budget is tight.

\begin{table}[t]
\caption{\footnotesize Refinement behavior on Video-MME. Panel (a) reports action frequencies by video length and overall. Counts are averaged over questions where the action occurs. Panel (b) reports refinement gains for overlapping question groups defined by whether each action occurred.}
\label{tab:refine-behavior}
\centering
\footnotesize
\renewcommand{\arraystretch}{0.65}
\setlength{\tabcolsep}{14pt}
\begin{clipbox}{0pt 0pt 0pt 0pt}
\begin{tabular}{@{}l rrrrr@{}}
\toprule
\rowcolor{cuepink!12}
\multicolumn{6}{@{}l}{(a) How often is each agentic refinement action used?} \\
Action & Short & Medium & Long & Overall & Count per question \\
\midrule
Keep    & 96\% & 92\% & 96\% & 94\% & 3.4 \\
Rewrite & 60\% & 74\% & 85\% & 73\% & 2.0 \\
Split   & 19\% & 30\% & 49\% & 33\% & 1.7 \\
Drop    & 63\% & 69\% & 80\% & 71\% & 3.9 \\
\addlinespace[1pt]
\rowcolor{cuepink!12}
\multicolumn{6}{@{}l}{(b) How much does refinement help each action group?} \\
\multicolumn{4}{@{}l}{Question group} & $\Delta$ at $k{=}8$ & $\Delta$ at $k{=}32$ \\
\midrule
\multicolumn{4}{@{}l}{Dropped a cue} & $+4.2$ & $+1.9$ \\
\multicolumn{4}{@{}l}{Rewrote a cue} & $+3.7$ & $+2.1$ \\
\multicolumn{4}{@{}l}{Split a cue}   & $+5.3$ & $+3.6$ \\
\bottomrule
\end{tabular}%
\end{clipbox}
\end{table}

\definecolor{cuerefine}{RGB}{255,242,204}
\paragraph{Cues retrieve evidence beyond the overview.}
Table~\ref{tab:explore} shows that initial cues match frames more strongly than the question in every duration split. Refinement further increases the relative gain over the overview, from $0.05$ to $0.07$ on short videos and from $0.43$ to $0.53$ on long videos. This widening gain suggests that refinement increasingly benefits exploration beyond the overview as videos grow longer. Split cues have the largest relative gain in every duration split, reaching $0.92$ on long videos. These gains support refinement's role in separating bundled visual elements for retrieval.

\begin{table}[t]
\caption{\footnotesize SigLIP2 score on Video-MME. For each cue set, we average per-cue maximum scores within each question and report the median across questions. The Question row reports the median maximum score for the question text. Candidate frames are 1-fps video frames (``Full video'') or the overview frames from \CD (``Overview''). Final cues comprise kept initial cues, rewritten cues, and split cues. Since Overview is a subset of Full video, the relative gain \text{Full}/\text{Overview}-1 is nonnegative. Larger gains indicate stronger matches beyond the overview.}
\label{tab:explore}
\centering
\footnotesize
\renewcommand{\arraystretch}{0.65}
\setlength{\tabcolsep}{18pt}
\begin{clipbox}{0pt 0pt 0pt 0pt}
\begin{tabular}{@{}ll ccc@{}}
\toprule
Retrieval text & Comparison & Short & Medium & Long \\
\midrule
Question & Full video & 0.34 & 0.79 & 0.77 \\
\addlinespace[1pt]
\rowcolor{cuecyan!11}
Initial cues & Full video & 0.89 & 0.95 & 0.96 \\
\rowcolor{cuecyan!11}
 & Overview & 0.85 & 0.81 & 0.67 \\
\rowcolor{cuecyan!11}
 & $\text{Full}/\text{Overview}-1$ & \emph{0.05} & \emph{0.17} & \emph{0.43} \\
\addlinespace[1pt]
\rowcolor{cuepurple!10}
Final cues & Full video & 0.93 & 0.98 & 0.98 \\
\rowcolor{cuepurple!10}
 & Overview & 0.87 & 0.82 & 0.64 \\
\rowcolor{cuepurple!10}
 & $\text{Full}/\text{Overview}-1$ & \emph{0.07} & \emph{0.20} & \emph{0.53} \\
\addlinespace[1pt]
\rowcolor{cuerefine!50}
\quad Rewritten cues & Full video & 0.98 & 0.99 & 1.00 \\
\rowcolor{cuerefine!50}
 & Overview & 0.93 & 0.88 & 0.69 \\
\rowcolor{cuerefine!50}
 & $\text{Full}/\text{Overview}-1$ & \emph{0.05} & \emph{0.13} & \emph{0.45} \\
\addlinespace[1pt]
\rowcolor{cuerefine!50}
\quad Split cues & Full video & 0.92 & 0.98 & 0.98 \\
\rowcolor{cuerefine!50}
 & Overview & 0.80 & 0.71 & 0.51 \\
\rowcolor{cuerefine!50}
 & $\text{Full}/\text{Overview}-1$ & \emph{0.15} & \emph{0.38} & \emph{0.92} \\
\bottomrule
\end{tabular}%
\end{clipbox}
\end{table}

\begin{table}[t]
\caption{\footnotesize Cue coverage on Video-MME. A frame is covered if any cue has a SigLIP2 score of at least $0.5$. Panel (a) reports median Jaccard overlap before and after refinement. Panel (b) reports QA accuracy (\%) with GPT-5.5 as the downstream VLM, given frames selected by \method, sampled outside \method's final cue coverage, or sampled uniformly. $\Delta$ is the overall accuracy difference from \method.}
\label{tab:coverage}
\centering
\footnotesize
\renewcommand{\arraystretch}{0.62}
\setlength{\tabcolsep}{14pt}
\begin{clipbox}{0pt 0pt 0pt 0pt}
\begin{tabular}{@{}c l rrrrr@{}}
\toprule
\multicolumn{2}{@{}l}{(a) Coverage overlap} & Short & Medium & Long & All & \\
\midrule
\rowcolor{cuepurple!10}
\multicolumn{2}{@{}l}{Initial cues vs. Final cues} & 0.89 & 0.74 & 0.65 & 0.76 & \\
\rowcolor{cuerefine!50}
\multicolumn{2}{@{}l}{Initial cues vs. Rewritten cues} & 0.49 & 0.41 & 0.36 & 0.39 & \\
\rowcolor{cuerefine!50}
\multicolumn{2}{@{}l}{Initial cues vs. Split cues} & 0.52 & 0.39 & 0.31 & 0.37 & \\
\addlinespace[1pt]
\midrule
\multicolumn{2}{@{}l}{(b) QA accuracy by frame source} & Short & Medium & Long & All & $\Delta$ \\
\midrule
\rowcolor{cuepurple!10}
\multirow{3}{*}{$k{=}8$} & \method & 85.56 & 77.56 & 74.78 & 79.30 & -- \\
 & Outside coverage & 72.78 & 69.00 & 67.22 & 69.67 & $-9.63$ \\
 & Uniform sampling & 79.89 & 71.67 & 69.56 & 73.70 & $-5.59$ \\
\midrule
\rowcolor{cuepurple!10}
\multirow{3}{*}{$k{=}16$} & \method & 87.33 & 79.11 & 75.44 & 80.63 & -- \\
 & Outside coverage & 75.00 & 73.11 & 67.56 & 71.89 & $-8.74$ \\
 & Uniform sampling & 83.67 & 76.11 & 71.56 & 77.11 & $-3.52$ \\
\midrule
\rowcolor{cuepurple!10}
\multirow{3}{*}{$k{=}32$} & \method & 89.89 & 83.11 & 76.56 & 83.19 & -- \\
 & Outside coverage & 78.33 & 76.33 & 72.56 & 75.74 & $-7.44$ \\
 & Uniform sampling & 89.00 & 80.00 & 75.22 & 81.41 & $-1.78$ \\
\bottomrule
\end{tabular}%
\end{clipbox}
\end{table}

\paragraph{Refinement redirects cue coverage.}
In Table~\ref{tab:coverage}(a), initial and final cue coverage have an overlap of only $0.76$ overall and $0.65$ on long videos. Rewritten and split cues have lower overlaps of $0.39$ and $0.37$ with their original coverage. Together with the stronger matches in Table~\ref{tab:explore}, these results indicate that refinement actively redirects cues to re-explore different parts of the video. This redirection becomes more pronounced as videos grow longer, with split-cue overlap falling from $0.52$ on short videos to $0.31$ on long videos.

\paragraph{\method explores useful regions.}
In Table~\ref{tab:coverage}(b), we select frames outside final cue coverage using farthest point sampling initialized by medoids. Accuracy is $7.44$ to $9.63$ points below \method\ and even below uniform sampling at every budget. Thus, \method\ explores question-relevant regions, while frames outside its coverage are less informative.

\begin{figure}[t]
    \centering
    \includegraphics[width=0.96\textwidth]{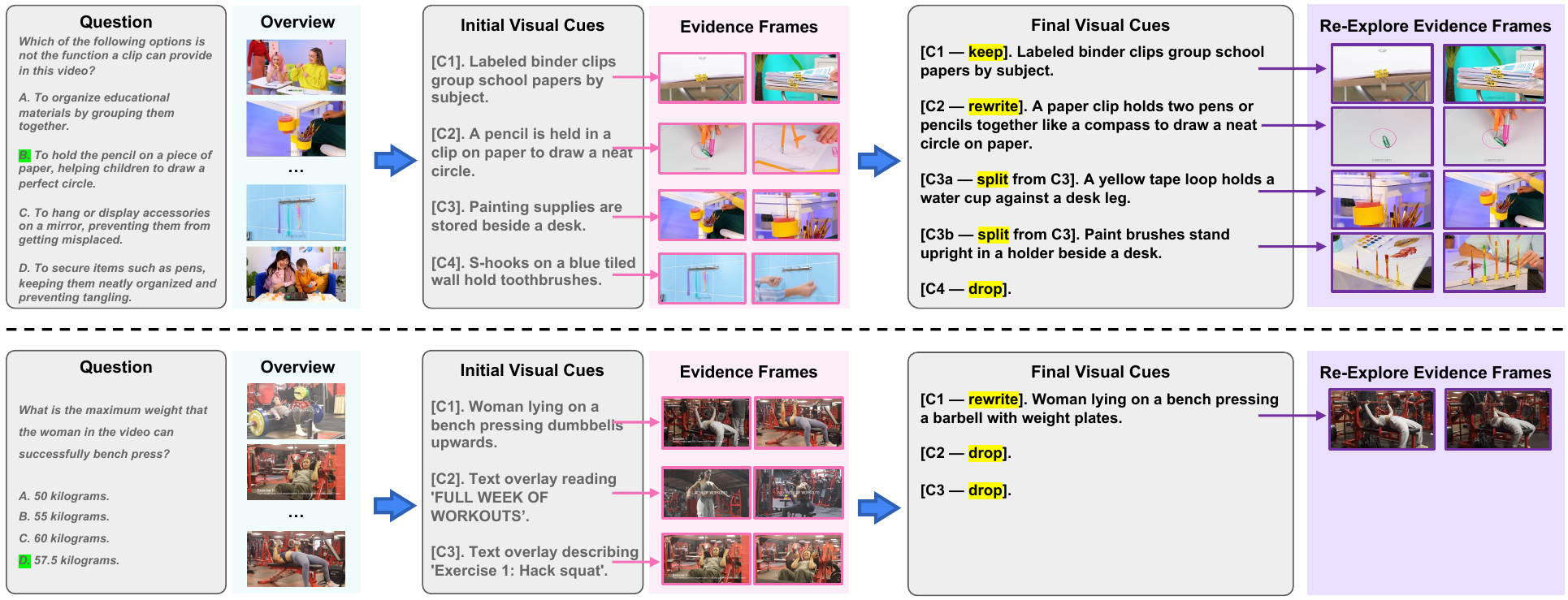}
    \vspace{-12pt}
    \caption{\small Qualitative illustration of \method: successful refinement (top) and failure (bottom). Correct options are green. We display selected overview frames and up to two evidence frames per cue per stage. The top panel is a curated mechanism illustration using real video frames; the bottom pairs recorded cues and actions with illustrative evidence frames.}
    \label{fig:qualitative}
    \vspace{-12pt}
\end{figure}

\subsection{Qualitative results}
\label{sec:qualitative}

\paragraph{Successful case.}
Figure~\ref{fig:qualitative} (top) illustrates each step of \method, starting with representative, diverse overview frames. The initial cues turn the visual elements in the answer options into distinct retrieval targets, allowing the first exploration to gather evidence frames for each candidate function. These evidence frames then benefit the agentic refinement to get the final cues. For example, C2 initially mentions ``a pencil is held in a clip on a paper'' based on option B, whereas its evidence reveals the clip-based construction. It triggers a rewrite as ``a paper clip holds two pens or pencils,'' which further unearths a new frame that shows more details before drawing circles, indicating that the clip is not used to hold the pencil to the paper. Similarly, splitting the broad C3 into a cup target and a paint-brush target prevents two different constructions from competing within one query and retrieves separate evidence for each. In contrast, C4 retrieves S-hooks holding toothbrushes, which cannot verify a function of a clip. Dropping it avoids allocating the final frame budget to this distractor. Keeping the already precise C1 preserves its relevant evidence. Together, these operations produce more specific and less redundant evidence for \CBA's final frame selection.

\paragraph{Failure case and limitations.}
The lower panel of Figure~\ref{fig:qualitative} exposes two limitations. First, open-ended cue discovery can introduce irrelevant targets: the wrong hack-squat cue is eventually dropped, but its retrieval and review still consume computation. Second, event-level information can be lost. Answering \emph{successfully} in this case requires evidence that a lift is completed, yet the VLM treats bench pressing as a single action and rewrites C1 only to identify a barbell, without separating the lowering, lifting, and completion stages. Re-exploration therefore retrieves the 60\,kg attempt but not its outcome, causing the downstream VLM to mistake a failed lift for a successful one; the correct maximum is 57.5\,kg. 
This failure exposes a limitation of the frame-wise embedding representation: since each frame is encoded independently, its similarity score does not represent the local temporal transition that determines whether the lift is completed. A temporal embedding model that represents each frame together with its neighboring frames could make such a short-lived transition appear as a coherent plateau rather than an isolated peak in our cue similarity wave, allowing our defined \emph{plateau} to retain evidence of the transition and its outcome.

\section{Conclusion}
We presented \method, a training-free method combining a reasoning VLM and an embedding model to explore long videos. Dynamic visual cues retrieve evidence that guides their refinement before frame budget allocation. Across three benchmarks, downstream VLMs, and frame budgets, \method\ establishes state-of-the-art results in all $27$ evaluated settings for which prior results are available. Ablations support its component designs, while behavioral analyses show useful retrieval beyond the overview and redirected exploration, particularly in longer videos. Overall, the results show that iterative, evidence-guided retrieval can sharpen underspecified cues, separate conflated targets, and suppress distractors before final selection. 
A promising direction is to incorporate temporal context into the embedding model to better capture event transitions and outcomes.

\clearpage
\subsection*{AI use statement}
Generative AI tools assisted with language revision and LaTeX typesetting. The authors determined the scientific content, verified all claims and reported values against the underlying experiments, and take responsibility for the final manuscript.

\subsection*{Reproducibility statement}
Section~\ref{sec:method} describes the method, and Section~\ref{sec:setup} specifies the evaluation protocol. Table~\ref{tab:hp} reports the reasoning VLM and hyperparameter selection. Appendix~\ref{app:impl} provides additional implementation details. Appendix~\ref{app:prompts} gives the prompt templates for \CD and the batched agentic refinement step of \CWGR.

\bibliography{references}
\bibliographystyle{iclr2027_conference}

\clearpage
\appendix
\section*{Appendix}
\setlength{\intextsep}{4pt plus 1pt minus 1pt}

\section{Additional implementation details}
\label{app:impl}
\paragraph{Image preprocessing and decoding.} Images supplied to SigLIP2 have a maximum edge of $384$ pixels. Qwen3.6-27B uses greedy decoding with temperature $0$.

\paragraph{\CD.} Overview segmentation uses a db4 discrete wavelet transform and retains its finest detail band. Boundary peaks are detected in the absolute reconstructed response, with height at least $\operatorname{mean}+0.5\operatorname{std}$, prominence at least $0.05$ times the response range, and separation of at least $\max(5,\lfloor 0.02n\rfloor)$ frames, where $n$ is the length of the signal currently being segmented, including recursive splits. Clip durations are constrained to between $4$ and $12$\,s. Overview slots are allocated approximately in proportion to each clip's candidate-frame count, with at least one slot per clip when the budget permits. We floor each clip's share of the total overview budget, enforce the one-slot minimum, and distribute leftover slots by fractional remainder; any excess is trimmed from clips with smaller weights that have more than one slot. Each clip's allocation is capped by its candidate count, and its medoid initializes within-clip farthest-point sampling. If the selected representatives underfill the budget, global farthest-point sampling adds unused frames, initialized with the selected representatives. If the medoids outnumber the overview budget, farthest-point sampling reduces the medoid set.

\paragraph{\CWGR.} The prominence base window extends $\delta{=}10$ frames on each side of a peak. Retained peak centers have a minimum separation of $d_p{=}3$ frames, with at most $n_p{=}5$ peaks per cue. Above-threshold intervals are merged when the next start index minus the previous end index is at most $2$, allowing at most one intervening below-threshold frame. A retained plateau spans at least $\ell_0{=}3$ frames and contributes $n_{\text{pl}}{=}3$ samples. Within each refinement batch, each cue with available evidence first receives one image. Each remaining image slot goes to the cue with the largest shortfall from its evidence-count-proportional share, up to its available evidence count. Evidence images shown to the reasoning VLM are arranged in temporal order within each cue and mapped to cue indices in the prompt.

\paragraph{\CBA.} Let $a_c$ be cue $c$'s retained evidence count and $r_c=k a_c/\sum_j a_j$ its target share of the total budget. If $k$ cannot cover all cues with evidence, the top-$k$ cues by rank receive one slot each. Otherwise, each nonempty cue starts with $k_c=1$; while slots remain, we increment the eligible cue with the largest $r_c-k_c$, subject to $k_c<a_c$, breaking ties by cue rank. If the total retained evidence count is at most $k$, all retained evidence frames are selected. If the retained evidence contains fewer than $k$ frames, farthest-point sampling in embedding space fills the remaining slots from unused candidate frames, initialized with the frames already selected.

\paragraph{Scores for cue analysis.} The analyses in Tables~\ref{tab:explore} and~\ref{tab:coverage} convert raw cosine similarities to $\sigma(\alpha s^c_t+\beta)$, where $\alpha>0$ and $\beta$ are SigLIP2's frozen scale and bias, and $\sigma$ is the sigmoid function.

\section{Full prompts}
\label{app:prompts}
Figures~\ref{fig:prompt-decompose},~\ref{fig:prompt-refine}, and~\ref{fig:prompt-qa} provide the prompt templates for \CD, the batched agentic refinement step of \CWGR, and downstream QA, respectively.

\begin{figure}[!ht]
\promptrole{System}
\begin{promptbox}
You are given some context about a video -- a question, an instruction, a
description target, or a query -- and a few sample frames from it. Write short
English search phrases ("cues"); each cue is later matched against the video's
frames to find where some visual thing appears. You only say what to look for;
you never answer the context.

Each cue:
- one self-contained sentence, < 25 words, describing what a single frame could
  show -- a subject, attribute, action, spatial relation, or on-screen text
  (quote the shortest useful fragment);
- specific enough to find by how a frame looks;
- distinct from the others;

When the context involves counting, ordering, before-after, or change-over-time,
write separate cues for distinct instances, events, or stages.

The attached frames are only a sparse sample of the video; if the context
needs something not visible in them, still write a cue for it.

Return one JSON object matching the schema, nothing else.
\end{promptbox}
\promptrole{User}
\begin{promptbox}
CONTEXT
"<question>"

CANDIDATE ANSWERS
A. <option A>   B. <option B>   ...

The attached frames are in time order.
Order the cues roughly by when each is likely to appear in the video.

Return exactly:
{ "visual_cue_list": [ {"cue": "<cue 1>"} ] }
\end{promptbox}
\caption{Prompt template for \CD. The reasoning VLM receives the task context and video overview and returns initial visual cues. The candidate-answer block is included only for multiple-choice inputs.}
\label{fig:prompt-decompose}
\end{figure}

\begin{figure}[!ht]
\promptrole{System}
\begin{promptbox}
You are given SEVERAL search phrases ("cues"), each with a few video frames that
look most like that cue.

For EACH cue, choose one action:
- drop:    the cue cannot help with the context.
- keep:    the cue is useful for the context; keep it unchanged.
- rewrite: the cue is useful but imprecise or retrieving the wrong thing; give
           one better cue, still faithful to the context.
- split:   the cue bundles two unrelated visual things; give two or more
           specific cues.

Return one JSON object with exactly one action PER cue, each tagged with that
cue's index, nothing else.
\end{promptbox}
\promptrole{User}
\begin{promptbox}
CONTEXT
"<question>"

CANDIDATE ANSWERS
A. <option A>   B. <option B>   ...

CUES (N total). The attached images are these cues' evidence frames concatenated
in cue order, in time order within each cue:
[0] "<cue 0>"  (images 1-4)
[1] "<cue 1>"  (images 5-7)
...

For EACH cue index return one action. Return exactly:
{ "reviews": [
  {"index": 0, "action": "keep"},
  {"index": 1, "action": "rewrite", "cue": "<rewritten cue>"},
  {"index": 2, "action": "split", "cues": ["<cue 1>", "<cue 2>", ...]},
  {"index": 3, "action": "drop"}
] }
\end{promptbox}
\caption{Prompt for batched agentic refinement in \CWGR. Each cue is paired with its own cue-wave evidence through an indexed image range. The reasoning VLM returns one keep, rewrite, split, or drop action per cue.}
\label{fig:prompt-refine}
\end{figure}

\begin{figure}[!ht]
\promptrole{System}
\begin{promptbox}
You are a helpful assistant.
\end{promptbox}

\promptrole{User: Video-MME}
\begin{promptbox}
Select the best answer to the following multiple-choice question based on the
video. Respond with only the letter (A, B, C, or D) of the correct option.
<question>
A. <option A>
B. <option B>
C. <option C>
D. <option D>

Answer with the option's letter from the given choices directly.
\end{promptbox}

\promptrole{User: LongVideoBench}
\begin{promptbox}
<question>
A. <option A>
B. <option B>
C. <option C>
D. <option D>
E. <option E, when present>
Answer with the option's letter from the given choices directly.
\end{promptbox}

\promptrole{User: MLVU}
\begin{promptbox}
<question and candidate options>
Only give the best option.
Best option: (
\end{promptbox}

\caption{Downstream QA prompts. All downstream VLMs receive the selected frames in temporal order with the corresponding dataset-specific user prompt.}
\label{fig:prompt-qa}
\end{figure}

\end{document}